\documentclass[runningheads]{llncs}
\usepackage{graphicx}
\usepackage{amsmath,amssymb} % define this before the line numbering.
\usepackage{color}
\usepackage[width=122mm,left=12mm,paperwidth=146mm,height=193mm,top=12mm,paperheight=217mm]{geometry}

\usepackage{subfig}
\usepackage{bm,array}
\usepackage{amsmath,amssymb} 
\usepackage{gensymb}
\usepackage{amssymb}
\usepackage{multirow}
\usepackage[utf8]{inputenc}
\usepackage[pagebackref=false,breaklinks=true,letterpaper=true,colorlinks,bookmarks=false]{hyperref}
\usepackage{booktabs}
\usepackage[font=small,labelfont=bf,tableposition=top]{caption}
\usepackage{adjustbox}

\DeclareCaptionLabelFormat{andtable}{#1~#2  \&  \tablename~\thetable}

\newcommand{\minisection}[1]{\noindent {\bf #1} }
\newcolumntype{C}{>{\centering\arraybackslash}p{5em}}

\begin{document}
\pagestyle{headings}
\mainmatter

\title{The CUDA LATCH Binary Descriptor:\\Because Sometimes Faster Means Better} % Replace with your title

\titlerunning{The CUDA LATCH Binary Descriptor}

\authorrunning{C. Parker, M. Daiter, K. Omar, G. Levi and T. Hassner}

\author{Christopher Parker$^{1}$, Matthew Daiter$^{2}$, Kareem Omar$^{3}$,\\Gil Levi$^{4}$ and Tal Hassner$^{5,6}$}
\institute{$^1$University of Oslo, Norway~~~~~$^2$Nomoko AG\\$^3$University of Alabama in Huntsville, AL, USA\\$^4$Tel Aviv University, Israel\\$^5$Information Sciences Institute, USC, CA, USA\\$^6$The Open University of Israel, Israel}%\\

\maketitle

\begin{abstract}
Accuracy, descriptor size, and the time required for extraction and matching are all important factors when selecting local image descriptors. To optimize over all these requirements, this paper presents a CUDA port for the recent Learned Arrangement of Three Patches (LATCH) binary descriptors to the GPU platform. The design of LATCH makes it well suited for GPU processing. Owing to its small size and binary nature, the GPU can further be used to efficiently match LATCH features. Taken together, this leads to breakneck descriptor extraction and matching speeds. We  evaluate the trade off between these speeds and the quality of results in a feature matching intensive application. To this end, we use our proposed CUDA LATCH (CLATCH) to recover structure from motion (SfM), comparing 3D reconstructions and speed using different representations. Our results show that CLATCH provides high quality 3D reconstructions at fractions of the time required by other representations, with little, if any, loss of reconstruction quality.
\end{abstract}

\section{Introduction}\label{sec:intro}
\begin{flushright}
{\em Quantity has a quality all its own}~~~~~~~~~~~\\
Thomas A. Callaghan Jr.
\end{flushright}
Local features and their descriptors play pivotal roles in many computer vision systems. As such, research on improving these methods has been immense. Over the years, this effort yielded progressively more accurate representations. These improvements were often demonstrated on standard benchmarks designed to measure the accuracy of descriptor matching in the presence of various image transformations and other confounding factors. It remains unclear, however, if the improved accuracy reported on these benchmarks reflects better, more useful representations when used in real world computer vision systems.

Take, for example, recent attempts to use deep learning for image feature representation (e.g.,~\cite{han2015matchnet,simo2015discriminative,zagoruyko2015learning}). There is no question that given sufficient training data and computational resources deep learning methods can achieve astonishing accuracy. Hence, using them to obtain local descriptors can result in better representations and by so doing impact a wide range of computer vision systems.

But using deep learning for feature description and matching does not come without a price: Most of these methods are computationally expensive and even with graphical processing units (GPU), are relatively slow. Even after extraction, their dimensions and floating point values makes them slow to match. Finally, they require substantial training data which can be difficult to provide.

These limitations should be contrasted with evidence that accuracy, though important, is not the only property worth considering when choosing descriptors. For example, simultaneous localization and mapping (SLAM) methods were shown to obtain better 3D reconstructions for the same computational effort by increasing the amount of feature points per keyframe~\cite{strasdat2012visual}. This suggests that computationally cheaper features are more desirable for these systems. Consequently, state of the art SLAM techniques~\cite{mur2015orb} use ORB~\cite{rublee2011orb} rather than more accurate but computationally expensive representations such as SIFT~\cite{lowe2004distinctive}: Doing so allows for a greater number of features to be extracted without compromising reconstruction accuracy. In fact, even classification systems appear to benefit from having more features over higher feature accuracy, as reported by~\cite{nowak2006sampling}.

One side effect to the success of deep learning is that the hardware enabling it -- GPU processors -- is now becoming standard on systems running computer vision applications, including even consumer cellphone devices. Beyond deep learning, these GPUs can also be used to accelerate extraction and matching of older, so-called {\em engineered} descriptors. These representations may not reach the same benchmark performances as deep learning techniques, but their extraction on the GPU offers a potential trade off between accuracy and run time. In particular, faster descriptor extraction and matching allows for more descriptors to be used and consequently better overall system performances. 

GPU accelerated features were considered in the past. We, however, focus on a particular binary descriptor: the {\em Learned Arrangement of Three Patches} (LATCH)~\cite{DBLP:conf/wacv/LeviH16}. It was recently shown to offer a compromise between the high accuracy, low speeds of floating point representations such as SIFT~\cite{lowe2004distinctive}, and the low accuracy, high speeds of binary descriptors (e.g.,~ORB~\cite{rublee2011orb}, BRIEF~\cite{calonder2010brief}). Beyond these properties, its design also happens to neatly fit GPU processing.

Our contributions are: {\bf (1)} We describe CLATCH, a CUDA port for LATCH, enabling descriptor extraction and matching directly on the GPU. {\bf (2)} We embed CLATCH in the OpenMVG library~\cite{openMVG}, along with a fast, GPU based Hamming distance, brute force descriptor matcher. Finally, {\bf (3)} we compare SfM 3D reconstructions on scenes from~\cite{DataSets3D} using SIFT, recent deep learning based representations and our CLATCH. These show that CLATCH reconstructions are comparable or even better than those obtained with other representations, yet CLATCH requires a fraction of the run time of its alternatives. Importantly, to promote reproducibility, the code used in this paper is publicly available from the project webpage:~\url{www.openu.ac.il/home/hassner/projects/LATCH}.

\section{Related work}\label{sec:related}
Due to their key role in many computer vision systems, local feature descriptors are extensively studied. A comprehensive survey is therefore outside the scope of this paper. Below we provide only a cursory overview of this topic. 

\minisection{Floating point representations.} For nearly two decades now, SIFT~\cite{lowe2004distinctive} is very likely the most widely used local image descriptor. It and the representations that followed (e.g., SURF~\cite{bay2006surf}) represent the region around an image pixel using a vector of typically 128 floating point values. This vector is often a histogram of measurements extracted from the image, most commonly various functions of the local intensity gradients. 

\minisection{Binary descriptors.} 
Despite the success of the older floating point representations, a prevailing problem was their extraction time and dimensionality (which, in turn, affected their storage and matching time). In response, binary descriptors were proposed as low dimensional, efficient alternative representations. These typically assign descriptor values by quick, pixel intensity comparisons. 

One of the first binary descriptors was the Binary Robust Independent Elementary Features (BRIEF)~\cite{calonder2010brief}, soon followed by the Oriented fast and Rotated BRIEF (ORB) descriptor~\cite{rublee2011orb} which added rotation invariance, the Binary Robust Invariant Scalable Keypoints (BRISK)~\cite{leutenegger2011brisk} which used a more effective pixel sampling pattern, and the Fast REtinA Keypoint descriptor (FREAK)~\cite{alahi2012freak} which sampled intensities using a pattern similar to the one found in human retinas. The Accelerated-KAZE (A-KAZE) was suggested in~\cite{Alcantarilla13bmvc}. It builds on the earlier Local Difference Binary (LDB) descriptor~\cite{yang2012ldb,yang2014ldb} by computing the binary descriptor values from mean image intensities over a range of patch sizes. The binary online learned descriptor (BOLD)~\cite{balntas2015bold} improve accuracy yet retain high processing speeds. Finally and very recently, the LATCH binary descriptors were proposed in~\cite{DBLP:conf/wacv/LeviH16}. We defer discussion of LATCH to Sec.~\ref{sec:latch}.

Hybrid binary/floating-point methods were also suggested. One example is LDA-Hash~\cite{strecha2012ldahash} which extracts SIFTs, projects them to a discriminative space and applies a threshold to obtain binary descriptors. DBRIEF~\cite{trzcinski2012efficient} instead uses patch intensities directly, BinBoost~\cite{lepetit2013boosting,trzcinski2013learning} learns a set of hash functions corresponding to each bit in the final descriptor and PR-proj~\cite{Simonyan14} uses learning and dimensionality reduction to produce compact binary representations. The computational effort required to extract these descriptors is similar to (if not greater than) floating point descriptors. The representations, however, are short binary vectors and so matching and storing them is relatively efficient.

\minisection{Computing local descriptors on the GPU.} Of course, we are not the first to propose porting local feature extraction to the GPU. To our knowledge, nearly all these efforts used the GPU to aid in extracting {\em floating point descriptors}, including GPU-SIFT (see, e.g.,~\cite{gpusift,Heymann2007SIFT,sinha2006gpu,Warn2009Accelerating}) and GPU-SURF~\cite{terriberry2008gpu}. These methods all used GPUs in portions of the extraction process. For example,~\cite{Warn2009Accelerating} used the GPU only to compute convolutions, all other stages performed on the CPU. In addition, and more importantly, the gain in performance reported by these methods are modest and do not approach the speeds of contemporary binary descriptors, let alone our CLATCH. 

Interestingly, the only available GPU {\em binary descriptor} is CUDA ORB, implemented by OpenCV~\cite{itseez2015opencv}. As we later discuss, due to the nature of GPU processing, the run time advantage of ORB over the more accurate LATCH descriptor when computed on the CPU, vanishes on the GPU.

\minisection{Deep features.} Following the remarkable success of deep learning in computer vision, it is no surprise that these methods are also being applied to feature point description. Convolutional Neural Networks (CNN) were used in a number of previous attempts to learn local descriptor representations~\cite{han2015matchnet,simo2015discriminative,zagoruyko2015learning,balntas2016pn}. 

In most cases, a Siamese deep network is trained with hinge loss~\cite{han2015matchnet,simo2015discriminative,zagoruyko2015learning}. The training set used in these cases consists of positive and negative labeled patch pairs. Metric learning is then used to improve matching. Finally,~\cite{balntas2016pn} proposed an efficient CNN design, bringing processing speeds down substantially. As we later show, their run time is still slower than our proposed approach.

\section{CUDA LATCH}\label{sec:latch}
\subsection{Preliminaries}
\minisection{The LATCH feature descriptor.}
LATCH was recently introduced in~\cite{DBLP:conf/wacv/LeviH16} and is available as part of the OpenCV library since ver. 3.0~\cite{itseez2015opencv}. Its design was inspired by the observation that {\em pure} binary descriptors such as BRIEF and ORB produce their values by comparing pairs of pixel intensities, a process which can be sensitive to local noise. To address this, these methods used various smoothing techniques before pixel values were compared. Smoothing, however, has the adverse effect of losing important high frequency image information. 

Rather than smoothing the image and then comparing single pixel values, LATCH computes its binary values by comparing pixel {\em patches}. The LATCH descriptor for image pixel $\mathbf{p}=(x,y)$ is computed by selecting $t=1..T$ patch triplets, one for each LATCH bit. For triplet $t$, three pixels are selected in the region around $\mathbf{p}$: an {\em anchor} pixel $\mathbf{p}_{t,A}$ and two {\em companion} pixels $\mathbf{p}_{t,1}$ and $\mathbf{p}_{t,2}$. The $k\times k$ pixel patches, $\mathbf{P}_{t,A}, \mathbf{P}_{t,1}$, and $\mathbf{P}_{t,2}$ centered on each of these three pixels are extracted. Finally, bit $t$ in the LATCH descriptor for $\mathbf{p}$ is set by comparing the Frobenious norm of the anchor to its two neighbors, as follows:
\begin{equation} 
LATCH(\mathbf{p},t) = \begin{cases} 1 & \mbox{if \ } ||\mathbf{P}_{t,A} - \mathbf{P}_{t,1}||^2_F > ||\mathbf{P}_{t,A} - \mathbf{P}_{t,2}||^2_F
\\ 0 & \mbox{otherwise} \end{cases}. \label{eq:latch}
\end{equation} 	
The triplets LATCH uses are fixed but are not arbitrary: Triplets are selected during training using the data set from~\cite{brown2011discriminative}, which contains same/not-same labeled image windows. Triplets were chosen by considering how well their bits correctly predicted the same/not-same labels over the entire training set. To prevent choosing correlated triplets, following~\cite{alahi2012freak,rublee2011orb}, triplets are skipped if their predictions are correlated with those of previously chosen triplets. 

In their work~\cite{DBLP:conf/wacv/LeviH16}, LATCH contained 512 bits (selected triplets) each one representing triplets of $7\times 7$ patches. At matching time, its computational requirements were obviously equal to those of any other 512 bit binary descriptor. Due to the use of patches and multiple Frobenious norms, extracting LATCH was slower than pure binary descriptors of the same size. Experiments reported in~\cite{DBLP:conf/wacv/LeviH16}, however, showed that the increase in extraction time was small. This was balanced by improved accuracy which bested existing binary descriptors, sometimes rivaling even larger floating point representations. 

\minisection{The GPU architecture and non-blocking programs.} Though the specific architectural designs of GPU processors changes from generation to generation, all have several multiprocessors. A CPU can launch {\em non-blocking} (parallel) GPU programs on these multiprocessors, referred to as {\em kernels}. That is, while a kernel is being executed on the GPU, the CPU is free to pursue other tasks and similarly, memory transfers to and from the GPU can take place without blocking either CPU or GPU. This property is extremely important when designing computer vision systems using the GPU: It implies that if the GPU extracts descriptors independently of the CPU, {\em the CPU is free to perform higher level processing}. Related to the SfM application considered here are optimizations for recovering transformations~\cite{hassner2014standard} and/or multiple view stereo for scene structure~\cite{furukawa2010accurate}.

Some previous attempts to port descriptors to the GPU used it only for parts of the descriptor extraction process, using the CPU for others and requiring multiple memory transfers between processors~\cite{sinha2006gpu,Warn2009Accelerating}. This at least partially explains why these attempts showed only modest run time improvements over their original, CPU implementations. As a design goal, we therefore limit the use of the CPU and any communications between it and the GPU when extracting and comparing our descriptors.

\minisection{Why LATCH?} LATCH was selected for following reasons.
\begin{itemize}
\item{\bf Memory access vs. computation.} The emphasis in GPU processing on raw arithmetic power results in memory access patterns often being the determining factor in performance rather than the actual computation. LATCH requires more processing than pure binary representations (e.g.,~\cite{calonder2010brief,rublee2011orb,leutenegger2011brisk,alahi2012freak,Alcantarilla13bmvc}) and therefore requires more CPU time to compute than they do. The memory transfer requirements of LATCH, however, are very similar to these other descriptors and hence it stands to gain more on the~GPU. 
\item{\bf Limited conditional branching.} As mentioned above, GPUs are optimized for processes which have few, if any, conditional branching; under these circumstances, modern GPUs are capable of up to 10 Tera-FLOPS. Most pure binary descriptors are therefore well suited for GPU processing, whereas porting more complex representations to the GPU is less trivial.
\item{\bf Binary string comparisons.} LATCH is a binary representation. Like other binary representations, it can be matched using fast Hamming distance comparisons. These can further be performed extremely fast on the GPU. 
\end{itemize}

Finally, as demonstrated in the tests reported by~\cite{DBLP:conf/wacv/LeviH16}, LATCH outperforms other binary descriptors making it ideally suited for our purposes. 

\subsection{Implementing LATCH with CUDA\protect\footnote{For brevity, only implementation highlights are provided. For more details, please see the code available from:~\url{www.openu.ac.il/home/hassner/projects/LATCH}.}}
We have ported the LATCH representation to CUDA 8, building on the original LATCH OpenCV C++ implementation. In all our evaluations, CLATCH representations were extracted from $64\times64$ pixel windows, using mini-patches of $8\times8$ pixels giving a 64-byte feature vector.

To minimize CPU processing, differently from~\cite{DBLP:conf/wacv/LeviH16}, we use the Features from Accelerated Segment Test (FAST)~\cite{rosten2006machine} feature detector. FAST is already available on the GPU as part of the OpenCV~\cite{itseez2015opencv} library. Given a detected oriented keypoint, $\mathbf{p} = (x,y,\theta)$ we extract LATCH from a $64\times 64$ intensities window around this point. This process is described next.

\minisection{Parallelizing LATCH on the GPU.} GPU kernels include several identical, concurrently-executing, non-interacting {\em blocks}. Each one consists of groups ({\em warps}) of 32 threads. In our implementation, a CLATCH kernel sequentially computes 16 descriptors per block. While the region of interest for one interest point is being processed, the next one is prefetched to pipeline the processing.

A single descriptor is extracted by multiple warps in each block. Each warp independently computes sixteen patch triplets, $[\mathbf{P}_{t,A},\mathbf{P}_{t,1},\mathbf{P}_{t,2}]$, four at a time, without any explicit synchronization during the main computation. All told, two blocks of 32 warps, each one containing 32 threads (total of 2048 threads) are processed at a time per multiprocessor. This coarse granularity was chosen to maximize performance across a variety of GPU architecture generations. 

\minisection{Memory optimization.} Given the FAST orientation for an image region, the rotated $64\times 64$ pixel rectangle is loaded into shared memory as an upright square of single-precision floats. We use texture memory accesses to efficiently load and process these values. Our implementation eliminates bank conflicts, with warp divergences or branches kept to a minimum. Thus, processing proceeds without {\texttt if} statements or communications between different warps. This is achieved by strided access patterns of patches and careful padding of shared memory, and is critical to CLATCH's high performance.

Specifically, patch comparisons are performed as follows. A warp simultaneously processes four triplets. Each thread (in a warp of 32 threads) performs two squared-distance comparisons per triplets in the F-norm of Eq.~\ref{eq:latch}. Then, fast warp shuffle operations are used to quickly sum the result from all pixel pairs in a novel, optimal manner. The original LATCH implementation used $7\times 7$ pixel patches. We use $8\times 8$ patches instead as this implies 64 values which can be handled concurrently with no extra computation costs. To further optimize this process, instruction level parallelism was exploited by manual loop unrolling and carefully arranging operations to prevent stalls due to data dependency.

\minisection{Weighing pixels in LATCH patches.} Each pixel within a patch can optionally be given a unique weight at no overhead. This is due to the GPU's emphasis on cheap fused-multiply-add operations. We use this property to simulate the original LATCH patch size of $7\times 7$ by setting the relevant weights to zero, obtaining the exact same representation as the original LATCH. Another potential use for this feature, not tested here, is applying Gaussian weights to patch pixels thereby better emphasizing similarity at the patch center vs. its outer pixels.

\section{SfM using CLATCH and OpenMVG}\label{sec:openmvgnovelties}
LATCH (and consequently CLATCH) were shown to be slightly less accurate than some of the more computationally heavy, floating point descriptors. It is not clear, however, how these differences in accuracy affect the overall accuracy and speed of an entire, descriptor-intensive computer vision system.

To this end, we test CLATCH vs. other descriptors on the challenging task of 3D SfM reconstruction. Our goal is to see how the final reconstruction and the time required to compute it are affected by the choice of descriptor. We use the OpenMVG, multiple view geometry library~\cite{openMVG}, modifying it to include self-contained CUDA streams and a GPU based, brute force Hamming matcher. These are detailed next.

\minisection{CUDA Integration.} The kernel launching mechanism employed by CUDA on its default stream disables concurrent launches of feature detection kernels. We therefore modified the CLATCH descriptor and matching code to exclusively operate off of self-contained streams. Doing so allowed the GPU to concurrently execute feature detection and description kernels across multiple images at once, as well as perform feature matching. 

\minisection{Descriptor matching on the GPU.} Our tests compare the use of our binary descriptors with existing floating point representations. In all cases, we used the GPU to compute the descriptor distances. Because CLATCH is a binary representation, Hamming distance is used to compute similarity of CLATCH descriptors. To this end, we developed our own GPU based Hamming distance brute force matcher and integrated it into OpenMVG. To provide a fair comparison, distances between floating point representations were computed using the standard OpenCV GPU based L2 distance routine. 

Each block of our Hamming-based brute force matching kernel processes half a probe descriptor per thread, though each descriptor is distributed throughout a half warp so that each thread holds parts of 16 probe descriptors. Gallery descriptors are alternatively prefetched into and processed from two shared memory buffers without intermediate synchronization. As the Hamming distance between each pair of probe and gallery descriptors is computed, partial results are distributed through a half warp. This calls for a simultaneous reduction of several independent variables, which minimizes the number of additions and warp shuffles to be performed. 

First, each thread halves the number of variables it must reduce by packing two variables into the lower and upper 16 bits of a 32 bit integer. Then, pairs of threads simultaneously exchange their packed variables in a warp shuffle, and sum the result with their original variable. This results in pairs of threads with variables holding identical values. The threads again pair off in the same manner, but exchange and sum a different variable. The same pairs of threads now have two variables with identical values, so the second of each pair of threads overwrites the first packed variable with the second, before each thread discards the second packed variable. This results in every thread in a warp having a unique value in the same variable, which allows efficient participation in subsequent warp shuffles until the reduction is complete. This novel method requires only 16 additions to compare 16 descriptor pairs, while the standard warp reduction pattern would require 80.

\section{Experiments}\label{sec:3dExperiments}
\minisection{Descriptor extraction run time comparison.} The CLATCH descriptor is identical to LATCH and so their accuracy on different benchmarks are the same. We therefore refer to the original paper for a comparison on standard benchmarks of LATCH and other representations~\cite{DBLP:conf/wacv/LeviH16}.

By using the GPU, CLATCH is much faster to extract. This is demonstrated in Table~\ref{tab:RunningTimes}, which provides a comparison of the run times reported for extracting many popular existing feature point descriptors compared to CLATCH. We report also the processor used to extract these representations and a price estimate for the processor in case of GPU based methods. 

Evident from the table is that even on affordable GPU hardware, extraction run times are orders of magnitude faster than standard representations on the CPU and even other GPU representations (the only exceptions are the far less accurate CUDA ORB and the floating point representation CUDA SURF). PN-Net~\cite{balntas2016pn} in particular, is designed to be a very fast deep learning based descriptor method, yet even with more expensive GPU hardware, it is more than an order of magnitude slower to extract than CLATCH. More importantly, all floating point representations, including CUDA SURF and PN-Net, require more time to match their bigger, real valued representations. 

\begin{table}[t]
	\centering
\begin{adjustbox}{max width=0.55\textwidth}
		\begin{tabular} {l @{~~~~~~~}  c c }
		\toprule
		Descriptor & Extraction $\mu S$ & GPU\\ \hline
		SIFT~\cite{lowe2004distinctive} & 3290 &  -\\ 
		SURF~\cite{bay2006surf} & 2110 &  -\\	
		CUDA SURF~\cite{bay2006surf}$^1$ & 0.9 &  GTX 970M (\$280 usd)\\	
		LDA-HASH~\cite{strecha2012ldahash} & 5030 & -\\ 
		LDA-DIF~\cite{strecha2012ldahash} & 4740 & -\\ 
		DBRIEF~\cite{trzcinski2012efficient} & 8750 & -\\ 
		BinBoost~\cite{lepetit2013boosting,trzcinski2013learning} & 3290 & -\\\hline	
		BRIEF~\cite{calonder2010brief} & 234 & -\\
		ORB~\cite{rublee2011orb} & 486 & -\\
		CUDA ORB~\cite{rublee2011orb}$^{1}$ & 0.5 & GTX 970M (\$280 usd)\\
		BRISK~\cite{leutenegger2011brisk} & 59& -\\
		FREAK~\cite{alahi2012freak} & 72 & -\\
		A-KAZE~\cite{Alcantarilla13bmvc} & 69& -\\ 
		LATCH~\cite{DBLP:conf/wacv/LeviH16} & 616 &  -\\ \hline
        DeepSiam~\cite{zagoruyko2015learning}$^{2,3}$ & 6580 & Titan (\$650 usd)\\
				MatchNet~\cite{han2015matchnet}$^4$& 575 & Titan X (\$1,000 usd) \\
        CNN3~\cite{simo2015discriminative}$^2$ & 760 & Titan Black (\$1,100 usd)\\ 
				PN-Net~\cite{balntas2016pn}$^2$ & 10 & Titan X (\$1,000 usd)\\ \hline 
		Our CLATCH & 0.5 & GTX 970M (\$280 usd)\\
		\bottomrule
\end{tabular}
\end{adjustbox}
		\caption{{\bf Run time analysis}. Mean time in microseconds for extracting a single local descriptor. For GPU descriptors we provide also the GPU models used to obtain these results and their estimated price. CPU results were all measured by~\cite{DBLP:conf/wacv/LeviH16} on their system. $^1$CUDA ports for SURF and ORB are implemented in OpenCV~\cite{itseez2015opencv}; their speeds were measured by us. $^2$Run times and hardware specs provided in the original publications. $^3$Time for extracting and matching a descriptor pair was reported as $\times$2 SIFT extraction time. $^4$Run time reported in~\cite{balntas2016pn}.}
		\label{tab:RunningTimes}
\end{table}

\minisection{SfM results.}
We use the incremental SfM pipeline implemented in OpenMVG, with its default values unchanged. We compared the following descriptors in our tests: SIFT~\cite{lowe2004distinctive}, often the standard in these applications, the deep learning based features, DeepSiam and DeepSiam2Stream from~\cite{zagoruyko2015learning}, the fast deep feature representations, PN-Net from~\cite{balntas2016pn} and our own CLATCH.

\begin{table}[th]
\begin{adjustbox}{max width=\textwidth}
\centering
\begin{tabular} {lcccccccc}
\toprule
 		& Avignon & Bouteville & Burgos & Cognac Garden & St. Jacques & Mirebeau & Murato & Poitiers\\ \hline
Number of images  & 11 & 26 & 9 & 12 & 20 & 22 & 43 & 33\\ \hline
\multicolumn{9}{l}{{\bf SfM Scene RMSE} (in pixels)} \\ \hline
SIFT~\cite{lowe2004distinctive}  & 0.475 & 0.405 & 0.495 & 0.438 & 0.498 & 0.478 & 0.533 & 0.690\\ 
DeepSiam~\cite{zagoruyko2015learning}& 0.533 & 0.489 & 0.422 & 0.566 & 0.535 & 0.489 & 0.533 & 0.547\\ 
DeepSiam2stream~\cite{zagoruyko2015learning}	& 0.505 & 0.457 & 0.419 & 0.536 & 0.522 & 0.459 & 0.496 & 0.529\\ 
PN-Net~\cite{balntas2016pn} 	& 0.538 & 0.462 & 0.493 & 0.554 & 0.536 & 0.482 & 0.531 & 0.533\\ 
Our CLATCH 	& 0.556 & 0.414 & 0.466 & 0.478 & 0.409 & 0.466 & 0.494 & 0.454\\  \hline
\multicolumn{9}{l}{{\bf Total time for descriptor extraction, matching and incremental SfM} (in seconds)} \\ \hline
SIFT~\cite{lowe2004distinctive}  & 174.30 & 454.13 & 143.728 & 155.61 & 296.431 & 401.64 & 958.778 & 1206.10\\ 
DeepSiam~\cite{zagoruyko2015learning} & 172.49 & 596.07 & 130.72 & 146.49 & 347.576 & 416.01 & 943.841 & 812.39\\ 
DeepSiam2stream~\cite{zagoruyko2015learning}	& 269.39 & 922.95 & 226.123 & 301.67 & 628.03 & 739.629 & 1750.535 & 1379.80\\ 
PN-Net~\cite{balntas2016pn} 	& 49.56 & 210.60 & 50.12 & 56.02 & 122.29 & 167.51 & 372.28 & 311.18\\ 
Our CLATCH 	& 18.91 & 69.07 & 15.907 & 19.089 & 27.877 & 47.534 & 86.377 & 61.868\\  \bottomrule
\end{tabular}
\end{adjustbox}
\caption{{\bf Quantitative SfM reconstructions.} Results on the eight scenes from~\cite{DataSets3D}, comparing various representations with our CLATCH. We report reprojection errors and the time required to extract, match and estimate shape for the various descriptors. All results measured on the same hardware. CLATCH run times are substantially faster than its alternatives despite similar qualitative results (see Fig.~\ref{fig:reconstruct}.)}
\label{tab:sfmResults}
\end{table}

\begin{figure}[ht]
\centering
\includegraphics[width=\textwidth,clip,trim = 0mm 0mm 0mm 0mm]{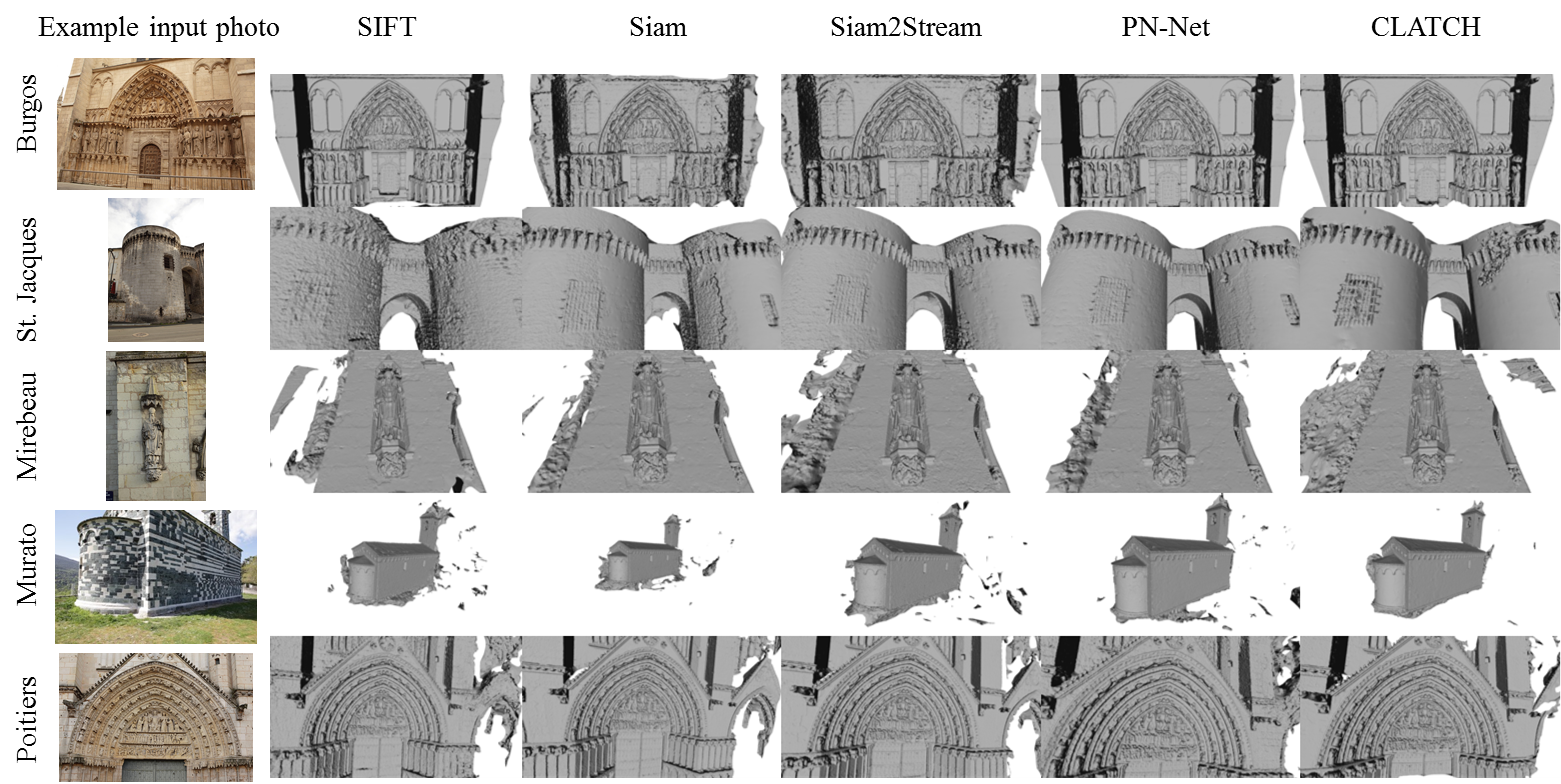}
\caption{{\bf Qualitative SfM reconstructions}. Results showing the output of the same SfM pipeline on five of the eights scenes from~\cite{DataSets3D}, comparing the use of SIFT~\cite{lowe2004distinctive}, Siam and Siam2Stream from~\cite{zagoruyko2015learning}, PN-Net~\cite{balntas2016pn} and our CLATCH. These results show only minor differences in output 3D shapes, despite the substantial difference in run time required for the different representations (see Table~\ref{tab:sfmResults}).}\label{fig:reconstruct}
\end{figure}

All descriptors used the CUDA FAST feature detector with the exception of SIFT which, for technical reasons, used its default DOG based detector. Following incremental SfM, point cloud Densification~\cite{barnes2009patchmatch}, Mesh Reconstruction~\cite{jancosek2014exploiting} and Mesh Refinement~\cite{vu2012high} were applied to produce the final reconstructions visualized in Fig.~\ref{fig:reconstruct}. 

Tests were performed on publicly available sets of high resolution photogrammetry images from~\cite{DataSets3D}, which include $5,616 \times 3,744$ (or $3,744 \times 5,616$) pixels in each image. Table~\ref{tab:sfmResults} summarizes these results, providing the final scene reprojection RMSE and the total time for descriptor extraction, matching and SfM reconstruction. All these tests were run on our GTX 1080 GPU.

Reconstruction run time is dominated by the brute force, nearest neighbor matcher. Hence, the gaps in run times between the different methods are smaller than those in Table~\ref{tab:RunningTimes}. Nevertheless, reconstructions with CLATCH required a fraction of the time for the runner up (PN-Net) and far less than the others. 

Reprojection RMSE, is low for all methods and is typically around half a pixel. Although these errors fluctuate between the different methods and scenes, these differences are often below 0.1 pixels. Considering the high resolutions of the input images, these differences are negligible. 

Finally, Fig.~\ref{fig:reconstruct} additionally provides qualitative results, showing rendered views of our reconstructions. Evident from the figure is that despite large differences in run time, qualitatively, the reconstructions appear very similar.

\section{Conclusions}\label{sec:conc} 
In descriptor matching intensive application, such as SfM, accuracy per descriptor is sometimes balanced by the speed required to extract and match the descriptors. Taking advantage of this, we present CLATCH, a CUDA port for the LATCH binary descriptor. Although CLATCH descriptor accuracy in standard benchmarks may fall slightly behind other representations, particularly recent deep learning based methods, they are far faster to extract and match. CLATCH thereby provides a fast and accurate alternative means for 3D reconstruction. 

From a technical point, an outcome of this work is openly available code for extremely fast feature extraction and matching and a pipeline for SfM allowing convenient interchange of feature descriptors, including deep methods. This implementation can be improved in many ways. For one thing, our use of the FAST detector~\cite{rosten2006machine} does not provide scale invariance. CLATCH, however, can easily be extracted at multiple scales, potentially improving its accuracy. Use of CLATCH in other applications where descriptors are extracted and matched in large quantities, is also a priority. One particularly appealing example is dense pixel matching~\cite{hassner2015dense,tau2016dense} where the CLATCH may be an alternative to methods such as PatchMatch~\cite{barnes2009patchmatch}, providing similar run times without compromising spatial smoothness.

\bibliographystyle{splncs03}%splncs}
%\bibliography{latch}

\end{document}